\documentclass[manuscript]{acmart}

\AtBeginDocument{%
  }

\setcopyright{acmlicensed}
\copyrightyear{2026}
\acmYear{2026}
\acmDOI{XXXXXXX.XXXXXXX}
\acmConference[Conference acronym 'XX]{Make sure to enter the correct
  conference title from your rights confirmation email}{April 13--17,
  2026}{Woodstock, NY}
\acmISBN{978-1-4503-XXXX-X/2018/06}

\usepackage{enumitem}
\setlist[itemize]{leftmargin=1.2em, labelsep=0.4em, itemsep=2pt, topsep=2pt}

\begin{document}

\title{XR\textsuperscript{3}: An E\underline{x}tended \underline{R}eality Platform for Social-physical Human-\underline{R}obot Interaction}

\author{Chao Wang}
\email{chao.wang@honda-ri.de}
\orcid{0000-0003-1913-2524}
\affiliation{%
  \institution{Honda Research Institute EU}
  \city{Offenbach am Main}
  \country{Germany}
}

\author{Anna Belardinelli}
\email{anna.belardinelli@honda-ri.de}
\orcid{0000-0002-0266-3305}
\affiliation{%
  \institution{Honda Research Institute EU}
  \city{Offenbach am Main}
  \country{Germany}
}

\author{Michael Gienger}
\orcid{0000-0001-8036-2519}
\affiliation{%
  \institution{Honda Research Institute EU}
  \city{Offenbach am Main}
  \country{Germany}
}

\renewcommand{\shortauthors}{Trovato et al.}

\begin{abstract}
Social-physical human–robot interaction (spHRI) is difficult to study: building and programming robots integrating multiple interaction modalities is costly and slow, while VR-based prototypes often lack physical contact capabilities, breaking the visuo-tactile expectations of the user.
We present XR\textsuperscript{3}, a co-located dual–VR-headset platform for HRI research in which a participant and an operator share the same physical space while experiencing different virtual embodiments. The participant sees an expressive virtual robot that interacts face-to-face in a shared virtual environment. In real time, the robot’s upper-body movements, head and gaze behaviors, and facial expressions are mapped from the operator’s tracked limbs and face signals. Since the operator is physically co-present and calibrated into the same coordinate frame, the operator can also touch the participant, enabling the participant to perceive robot touch synchronized with the visual perception of the robot’s hands on their hands: the operator's finger and hand motion is mapped to the robot avatar using inverse kinematics to support precise contact. Beyond faithful motion retargeting for hand and head motion, our XR\textsuperscript{3} system supports social retargeting of multiple nonverbal cues 
, which can be experimentally varied and investigated while keeping the physical interaction constant. We detail the system design and calibration, and demonstrate how the platform can be used for experimentation and data collection in a touch-based Wizard-of-Oz HRI study, thus illustrating how XR\textsuperscript{3} lowers barriers for rapidly prototyping and evaluating embodied, contact-based robot behaviors.
\end{abstract}


\begin{teaserfigure}
  \includegraphics[width=\textwidth]{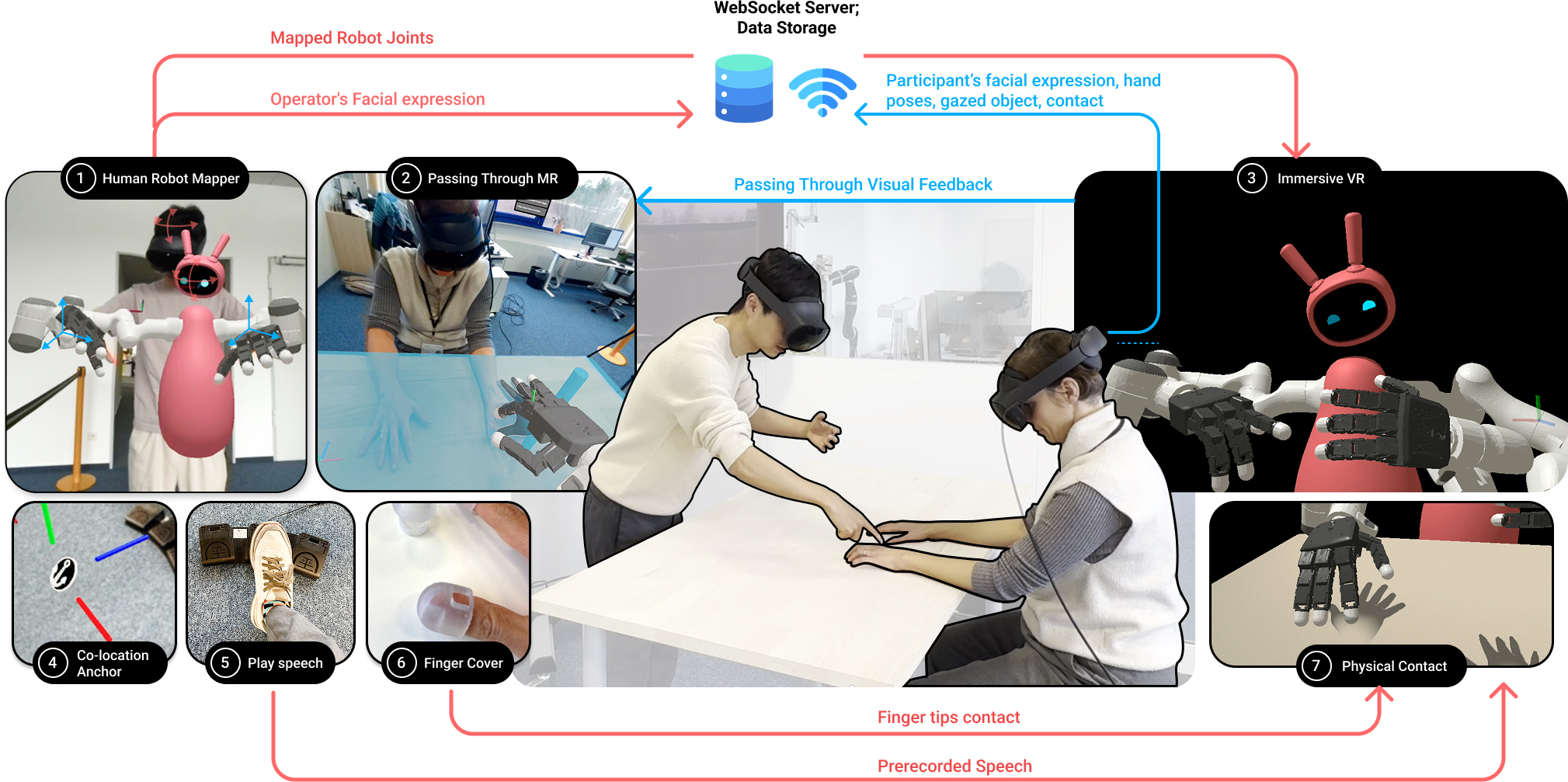}
  \caption{XR\textsuperscript{3}: a co-located dual-headset platform for contact-rich interactions. 
A participant (placed in an immersive VR scene) interacts face-to-face with a virtual robot, while an operator, co-located in the same space, but using passthrough MR, drives the robot via real-time motion and facial retargeting. 
A shared spatial anchor aligns both headsets in one coordinate frame, enabling realistic visuo-tactile stimulation of the participant. 
A foot pedal triggers pre-recorded speech for repeatable, hands-free Wizard-of-Oz control of verbal behavior. 3D-printed fingertip covers  match the robot fingertip geometry and appearance during contact (passive haptics).}
  \Description{}
  \label{fig:teaser}
\end{teaserfigure}

\received{20 February 2007}
\received[revised]{12 March 2009}
\received[accepted]{5 June 2009}

\maketitle

\section{Introduction}
Design, prototyping, and user testing should ideally go hand in hand in human-robot interaction; yet, this is rarely the case. In particular, in the context of spHRI, robots need to share a space and act in tight contact with humans, thus both psychological and physical engagement need to be supported \cite{Farajtabar2024}. That is, even if the robot's main purpose is physical support, this should be accompanied by a transparent and legible social behavior to improve acceptance and user experience \cite{Nguyen2018,Seifi2024}. Social robotics has long investigated which cues and in which combinations help in perceiving a robot as a sociable companion \cite{Hirano2018, Saunderson2019}. Nevertheless, how to realize the multimodal, flexible, and context-appropriate social behavior that makes physical robots acceptable and trustworthy, in our homes or -- even closer -- in our peripersonal space, remains an open question.  
While robotics has made huge progress in generating smooth, lifelike motions for physical tasks, it remains challenging to display appropriate and safe behavior during human-robot interaction. The robot's social and functional behavior should be co-designed along with its hardware and iteratively assessed with users, but this is often possible only once a full hardware prototype and behavior implementation is available. However, experimenting with real and fully autonomous robots is often complicated and time-consuming. Even if focusing just on designing an interaction with a commercial platform, comparing it to another embodiment and showing the influence of a specific appearance in the face of the same behavior is not always straightforward. 
Wizard-of-Oz studies \cite[WOZ]{riek2012wizard} help with prototyping but typically accommodate limited control of the robot (e.g., verbal interaction), while other performative approaches contemplating a human proxy role-playing the robot \cite{rozendaal2024enacting} are more suitable for ideation and participatory design than for experimental testing.
For these reasons, appearance and social aspects related to touch-based interactions are often investigated with pictures or videos \cite{Fitter2016,Mazursky2022,Law2021}, with participants left to imagine how the interaction would feel in first person. In this sense, Virtual Reality (VR) has been recognized for some time as a critical tool to assess numerous dimensions of HRI \cite{Villani2018,williams2018virtual}. Still, physical interactions are typically hard to realize in VR, since a simulated robot cannot make real contact with the participant. 

To deal with such limitations, we propose a framework combining the immersive experience offered by VR with the safety and behavioral richness of WOZ studies and the experimental controllability of MR, thus realizing a versatile XR system. We particularly focus on multimodal social (cue) retargeting, a method relying on a VR-MR setup to display, capture, and validate robot social behavior during a realistic interaction, allowing safe and user-centric spHRI already at the design stage. 
For our research purposes, we are interested in face-to-face interaction with an anthropomorphic robot. To account for realistic kinematics comparable to our actual physical manipulators (2 Franka Emika Panda arms\footnote{https://franka.de/franka-research-3} endowed with Allegro hands\footnote{https://www.allegrohand.com/}) while allowing design exploration for the robot appearance beyond human-like features, our exemplary virtual embodiment for social retargeting features a roundish trunk and a cartoon-like head with a screen face and actuated ears (see Figures \ref{fig:teaser}, \ref{fig:Mapping}). XR\textsuperscript{3} is not tailored to this embodiment; however, to detail its workings, we will refer to the specifics of this embodiment.




\section{Dual VR co-location system}
\label{sec:system}

XR\textsuperscript{3} is a co-located dual-headset platform for allowing touch-based HRI studies (Figure~\ref{fig:teaser}). 
A participant and a hidden operator share the same physical space but see different views: the participant is immersed in VR, facing a 3D robot avatar, while the operator uses passthrough Mixed Reality (MR) to control the robot and can physically interact with the participant to provide synchronized visuo-haptic stimulation.

\subsection{System features}
\label{subsec:features}

\subsubsection{Co-location and asymmetric views}
\label{subsec:colocation}
\begin{itemize} 
    \item \textbf{Co-location via spatial anchor.} Two VR devices are spatially calibrated and share a common coordinate system (e.g., via a Meta spatial anchor,  see panel~4 in Figure~\ref{fig:teaser}). This avoids marker-based calibration (e.g. with fiducial markers like ArUco) and supports quick setup in different rooms.
    \item \textbf{Runtime placement adjustment.} During setup, the operator can further adjust the robot’s virtual placement (e.g., small translation/yaw offsets) through a GUI to match the physical arrangement (e.g., across a table).
    \item \textbf{Asymmetric rendering for control and immersion.} Through passthrough MR, the operator sees the real environment with the robot overlaid on their body, enabling continuous awareness of the participant’s posture and proximity while maintaining alignment for safe touch (panel~2 in Figure~\ref{fig:teaser}). The participant experiences a fully virtual environment and sees the robot and their own virtual hands (panel~3 in Figure~\ref{fig:teaser}).
\end{itemize}

\subsubsection{Communication, logging, and touch aids}
\label{subsubsec:comm_log_touch}
\begin{itemize}
    \item \textbf{Realtime streaming.} The robot avatar's motion data from the operator headset---including arm/hand motion, head orientation, and face expressions (Section~\ref{subsec:mapping})---is streamed to the participant's headset via a WebSocket server.
    \item \textbf{Synchronized logging.} In parallel, the server logs multimodal data for analysis: robot pose/control streams, both operator and participant face blendshapes and hand joints, and participant's interaction events such as gazed robot body parts (e.g., head/hands/torso) and hand collisions with robot hand.
    \item \textbf{Pre-recorded speech triggering.} To keep the operator's hands free for the physical interaction, XR\textsuperscript{3} integrates a foot pedal to trigger pre-recorded utterances (panel~5 in Figure~\ref{fig:teaser}). The pedal provides three buttons and supports single- and double-press bindings, enabling quick navigation (e.g., previous/next clip) and branching between alternative replies.
    \item \textbf{Touch alignment aids.} The operator can physically touch the co-located participant while the participant sees the robot touching hand. To better match the robot fingertip geometry and materials during contact, the operator can wear 3D-printed fingertip covers shaped like the virtual robot's fingertips (panel~6 in Figure~\ref{fig:teaser}).
\end{itemize}

\subsection{Physical and social motion retargeting}
\label{subsec:mapping}

\begin{figure*}[!t]
    \centering
    \includegraphics[width=1\linewidth]{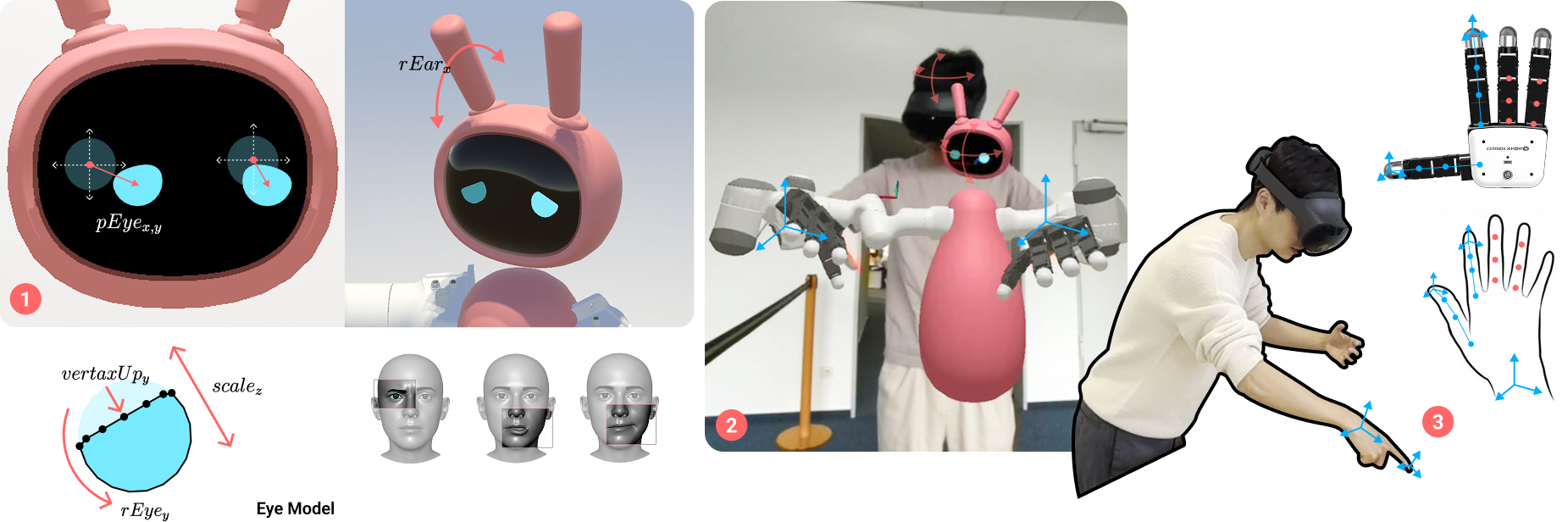} 
    \caption{Human-to-robot mapping in XR\textsuperscript{3}. 
    (1) Face signals (selected blendshapes and gaze) control robot ear rotation and screen-eye shape/pose. 
    (2) The operator's head orientation is retargeted to the robot's head; the operator's 6d hand pose provides are projected on the robot's end effectors, and the robot arm joint motion is resolved by Inverse Kinematics (IK). 
    (3) Thumb and index fingertips follow the operator's fingertip positions via fingertip IK; the remaining finger joints use direct joint-angle mapping.}
    \label{fig:Mapping}
\end{figure*}

XR\textsuperscript{3} retargets the operator’s tracked head, hands, and face signals to the robot avatar in real time (Figure~\ref{fig:Mapping}). 
The mapping has two goals: (1) enable expressive and legible nonverbal behaviors (gaze, facial expressions, head orientation), and (2) support \emph{touch alignment} by making the robot’s visible hand/fingertip contacts coincide with the operator’s physical contacts with the participant.

\subsubsection{Retargeting pipeline}
\label{subsubsec:retarget}
\begin{itemize}
    \item \textbf{Face and gaze $\rightarrow$ screen-face controls.} 
    The employed VR device (Quest Pro) provides face tracking that outputs 70 muscle blendshapes as values in $[0,1]$.\footnote{\url{https://developers.meta.com/horizon/documentation/unity/move-face-tracking}}
    We select seven blendshapes that we found to most affect perceived emotion in our screen-face embodiment:
    \emph{eye closedness} $C_{\text{eye}}$, \emph{lip-corner dimple} $D_{\text{lip}}$, \emph{brow lower} $H_{\text{brow}}$ (left/right), and \emph{chin raise} $H_{\text{chin}}$; the headset also provides gaze direction $(\theta_x,\theta_y)$.
    Our robot face uses two stylized ``eye'' cylinders rendered on a screen; their shape and pose are controlled by parameters including eyelid-profile vertices $\text{vertexUp}_y$ and $\text{vertexLow}_y$, eye rotation $r_{\text{Eye}y}$, eye depth scale $s_{\text{Eye}z}$, ear rotation $r_{\text{Ear}x}$ (one DoF per ear), and on-screen eye position $(p_{\text{Eye}x},p_{\text{Eye}y})$ (Figure~\ref{fig:Mapping}, step~1).
    We use a simple monotonic mapping that (i) closes/compresses the eyes with $C_{\text{eye}}$, (ii) modulates eye/ear ``attitude'' with combined brow/chin activity, and (iii) drives on-screen eye position from gaze with normalization and clamping; full equations are provided in Appendix~\ref{app:face-mapping}.

    \item \textbf{Head orientation retargeting.}
    The headset provides a 6-DoF head pose. Because the robot base is fixed in the scene, we retarget only the operator’s head \emph{orientation} to the robot head while discarding translation to keep the robot rooted.

    \item \textbf{Hands $\rightarrow$ robot arms via IK for touch alignment.}
    The headset provides articulated hand tracking (skeletal joint transforms). Our goal is not only plausible robot motion but also touch alignment: when the participant sees the robot end-effector contacting their hand/arm, they should simultaneously feel contact at the corresponding location.
    We achieve this by overlaying the robot avatar onto the co-located physical interaction space (Section~\ref{subsec:colocation}) and solving inverse kinematics (IK) chains to match tracked human hand targets (Figure~\ref{fig:Mapping}, steps~2--3).
    Specifically, we use three IK chains:
    (i) \textbf{arm-to-palm IK} from the robot arm's base link to the robot's palm,
    (ii) \textbf{thumb fingertip IK} from thumb base to thumb tip using the tracked thumb tip pose, and
    (iii) \textbf{index fingertip IK} from index base to index tip using the tracked index tip pose.
    This prioritizes precise alignment at the palm and two salient fingertips that frequently initiate and regulate contact, enabling more convincing ``robot touch'' during fingertip interactions.

    \item \textbf{Remaining fingers via direct joint-angle mapping.}
    For the remaining fingers (middle, ring, little), we perform direct joint-angle mapping: each robot finger joint follows the corresponding human finger joint rotation (the little finger is ignored since the robot hand only has four fingers). 
    We do not solve IK for all fingertips because the used hand is substantially larger than a human hand, making simultaneous absolute alignment of all five fingertip poses difficult without introducing large errors or unnatural postures. In practice, full-fingertip IK tended to destabilize contact alignment during dexterous motion; therefore, we prioritize thumb and index fingertips.

    \item \textbf{Contact realism with fingertip covers.}
    To further increase realism during fingertip touch, the operator can wear 3D-printed fingertip covers shaped to match the robot's fingertip geometry (panel~6 in Figure~\ref{fig:teaser}), helping the \emph{felt} contact to better match the \emph{seen} robot fingertip (e.g., taps and light presses).
\end{itemize}

\begin{figure}[t]
\centering
\includegraphics[width=0.9\linewidth]{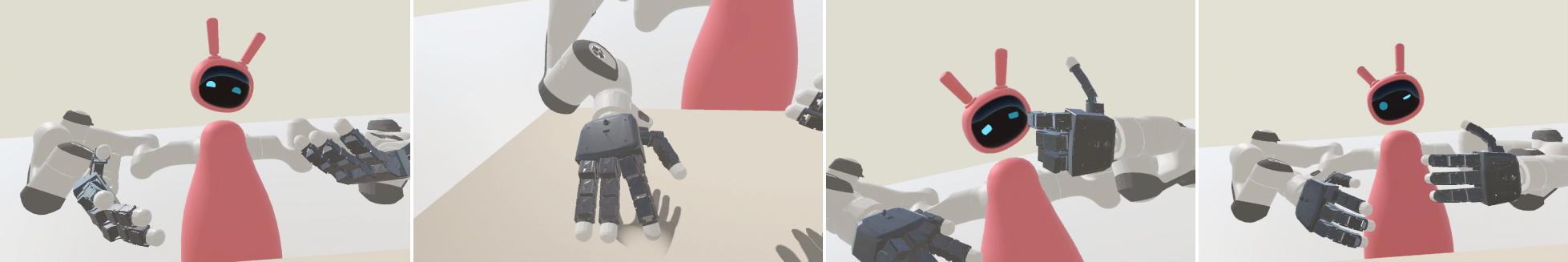}
\caption{The behavior of the virtual robot from the participant's point of view during the experiment. From the left: the robot introduces itself, draws a pattern on the hand of the participant, acknowledges their right guess, concludes the interaction.}
\label{fig:scenario}
\end{figure}

\section{Conducting user studies}

To test and validate the proposed framework, we are conducting a pilot study manipulating factors that are specifically of interest to our research on socio-physical HRI, but that can also showcase how the XR\textsuperscript{3} system can be used for prototyping and conducting user studies in a convenient balance of experimental controllability and interaction realism.
\subsection{Exemplary experimental interaction} 
The study is meant to investigate how the level of social retargeting can affect user experience and trust in the context of robot-initiated contact. 
Here, we want to show that the XR\textsuperscript{3} system can enhance the sense of being in the presence of the robot and provide a more realistic experience of being touched by it, while still keeping participants safe.
In a Wizard-Of-Oz experiment we hypothesize that the experience of (functional/affective) touch by the robot can be improved the more human-like the robot's social behavior, intended as a combination of retargeted non-verbal cues (3 levels of non-verbal expressivity: 1) just \textit{head} motion; 2) \textit{head and eye} movements; 3) \textit{head motion, eye movements, and face expressions}). On the other hand, such experience can be influenced by the context and function of the administered touch (2 levels of context: \textit{functional} vs. \textit{playful}).
Participants will participate in either context and are informed that they will interact with a robot in VR across three trials, each time the robot has different expressive capabilities. In each interaction, the robot first introduces itself as a nurse (\textit{functional} context) or a robot companion (\textit{playful} context), asks the participant to turn up the hand it briefly taps, explains it will draw a shape for the participant to guess, draws it on the participant's palm and acknowledges the answer, and finally concludes the interaction (see Figure \ref{fig:scenario} for salient moments).
Throughout the study, the wizard will be played by the same operator to minimize differences in stimulus administration across participants. The wizard will follow a precise script, with consistent movements and face expressions, thoroughly rehearsed, across all conditions. That is, across expressivity conditions, the wizard performs always the same social and physical behavior, but only the condition-related cues are retargeted and displayed by the robot in VR. Speech lines are always the same within each context scenario and delivered as explained in Section \ref{subsec:features}. Essentially, the wizard is solely responsible for turn-taking, triggering a new utterance only when the participant is ready, and behaving consistently in every trial, independent of context and expressivity levels.
\begin{figure}[t]
\centering
\includegraphics[height=3.25cm, keepaspectratio]{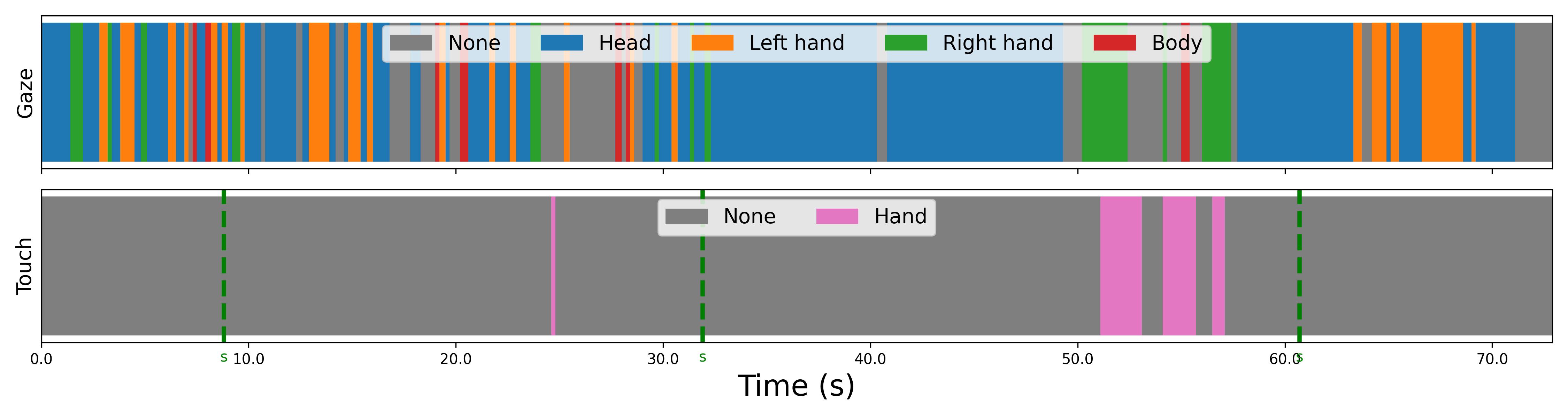}
\includegraphics[height=3.25cm, keepaspectratio]{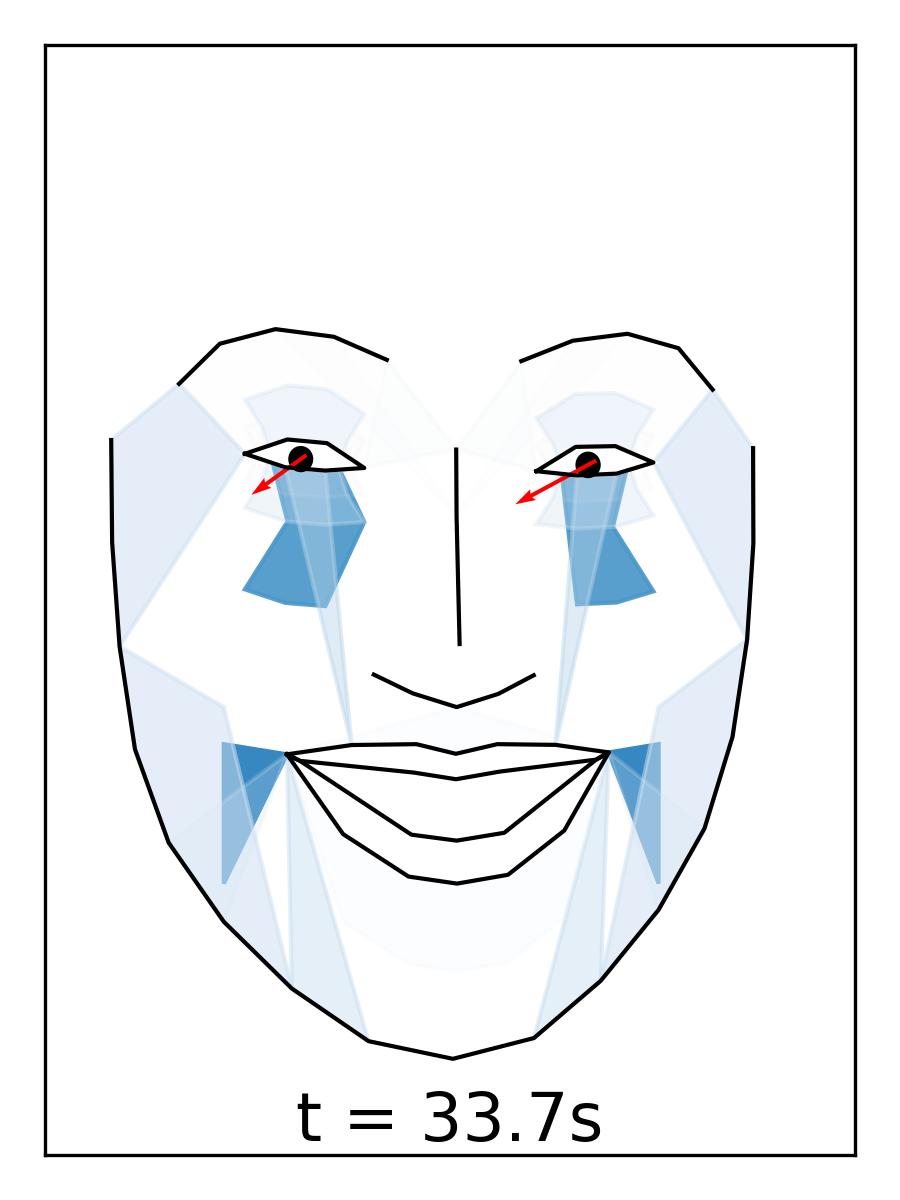}
\caption{Plots of participant gaze and touch events (left) as collected along one trial. The vertical dashed lines on the bottom plot denote the start of speech utterances. Aligned time series are collected for 25 AUs, providing face expressions related to such events. For example after the first tap on the hand (t=$\sim25s$), high values of AUs typically related to happiness/smiling  are detected (cheeck raiser and lip corner puller, see visualization on the right, \cite{cheong2023py}), and the person shifts their gaze between the head and the hands of the robot. 
}
\label{fig:data_visualization}
\end{figure}

\subsection{Data Collection}

By means of the XR\textsuperscript{3} framework, we can collect timestamped data streams from both headsets during each interaction. In particular, for the participant, at any time point, we collect which part of the robot body (head, trunk, hands) they are focusing on (intersection of the gaze vector with a robot collider). For the robot/operator, we collect contact events with the participant by detecting collisions between the robot's hand and the participant's hand colliders. We further collect all blendshapes from both headsets and compute action units (AUs) according to the Facial Action Coding System (\cite{Ekman1978}). Such features have been recently related to user understanding and trust in robotic partners \cite{Green2025,Kontogiorgos2025}. We further collect speech events considering pedal key presses. Pose data from both actors (heads, hands, or other joints of interest) can also be acquired. All this data can be used to relate specific experimental events in time (robot speech, robot physical contact)  to behavioral reactions from the participants (gaze, face expressions) as proxies of attention, acceptance/trust, or other attitudes toward the robot (see Figure \ref{fig:data_visualization}).  In aggregate form, such behavioral measurements can be compared across conditions and correlated with subjective measurements, such as questionnaires administered at the end of each interaction, still providing a more granular picture of the dynamic nature of the investigated constructs.   

\section{Conclusion}
The proposed XR\textsuperscript{3} framework combines motion and social retargeting to allow an operator to enact a robot with realistic behaviors, appropriately modulated by human social cognition. It can be conveniently used to quickly prototype and validate even social behaviors entailing physical contact, in a safe and user-centric way. Multimodal data from dyadic interactions can be collected in sync  and in anonymized form and used for various research purposes: data from the wizard can be used for training a robot with appropriate behaviour, in terms of timed gestures, expressions, contacts, w.r.t. the current situation (e.g., serious/happy/empathic/…), data from the user can be used to evaluate behavioral correlates of user experience (e.g., hand motion data or face expressions/face action units as proxies of stress, distrust, discomfort) in connection with specific interaction events and in relation to post-interaction subjective scales.


\bibliographystyle{ACM-Reference-Format}
\bibliography{references,VR2VR}

\appendix
\section{Facial-expression mapping equations}
\label{app:face-mapping}

We map a subset of Quest Pro blendshape coefficients and gaze direction to the robot screen-face parameters used in Figure~\ref{fig:Mapping} (step~1).
Let $\text{vertexUp}_y$, $\text{vertexLow}_y$, $r_{\text{Eye}y}$, $s_{\text{Eye}z}$, $r_{\text{Ear}x}$, and $(p_{\text{Eye}x},p_{\text{Eye}y})$ denote the controllable degrees of freedom.
The mapping is:
\begin{align*}
\text{vertexLow}_y &= \min\!\bigl(D_{\text{lip}},\;\text{vertexLow}_y\bigr),\\
\text{vertexUp}_y  &= \max\!\bigl(-\tfrac{H_{\text{chin}}+H_{\text{brow}}}{2},\;\text{vertaxUp}_y\bigr),\\
r_{\text{Eye}y}     &= (H_{\text{chin}}+H_{\text{brow}})\,\pi/6,\\
r_{\text{Ear}x}     &= \pi/2\,(-H_{\text{chin}}+H_{\text{brow}}),\\
s_{\text{Eye}z}     &= 1-0.9\,C_{\text{eye}},\\
(p_{\text{Eye}x},p_{\text{Eye}y}) &= \operatorname{clamp}\!\bigl(-(\theta_x,\theta_y)/\theta_{\max},\,-1,1\bigr).
\end{align*}
Here, $\theta_{\max}$ is the maximum gaze angle used for normalization, and $\operatorname{clamp}(\cdot)$ limits each component to $[-1,1]$.

\end{document}